\documentclass[conference]{IEEEtran}
\IEEEoverridecommandlockouts
\usepackage[utf8]{inputenc}
\usepackage{amsmath,amssymb,amsfonts}
\usepackage{graphicx}
\usepackage{booktabs}
\usepackage{hyperref}
\usepackage{microtype}
\usepackage{subcaption}
\usepackage{cite}
\usepackage{url}
\usepackage{algorithm}
\usepackage{algpseudocode}
\usepackage{enumitem}
\setlist{noitemsep,topsep=2pt,parsep=1pt,partopsep=1pt}

\title{\textbf{EngramRAG: Dynamic Usage-Weighted Topology and Synaptic Consolidation for Multi-Hop Agentic Memory}}

\author{
  \IEEEauthorblockN{Bhavyateja Potineni, Lohit Giri, Anu Jain, Vadim Kutsyy, and Rajasekhar Pentakota}
  \IEEEauthorblockA{\textit{Independent Researchers}}
}

\begin{document}

\maketitle

\begin{abstract}
As autonomous Large Language Model (LLM) agents are deployed across sustained, multi-session environments, conventional memory architectures suffer from three fundamental pathologies: (1) \textit{Associative Blindness}, wherein flat dense vector indexes fail to traverse multi-hop relational dependencies across distant interaction sessions; (2) \textit{Scaffolding Amnesia}, where standard wall-clock temporal decay rules aggressively erase foundational user identity and architectural constraints; and (3) \textit{Static Topology Stagnation}, where existing graph-augmented retrieval systems treat network structure as immutable, ignoring real-time usage dynamics and task utility. In this work, we propose \textbf{EngramRAG}, a hierarchical hybrid memory architecture for autonomous agents grounded in Complementary Learning Systems (CLS) principles. EngramRAG partitions memory operations into a low-latency streaming ingestion and retrieval reflex (``Waking State'') and an asynchronous background synaptic consolidation cycle (``Dreaming State''). EngramRAG introduces Usage-Modulated Personalized PageRank (U-PPR), a spreading activation algorithm wherein transition probabilities adapt via activity-dependent Hebbian plasticity to dynamically promote persistent foundational entities into high-centrality Epistemic Macro-Hubs; Consolidation-Activated Topology Decay (CATD), a selective forgetting rule executed strictly off the read path that scales retention half-life by topological load-bearing weight rather than wall-clock recency, protected by a cold-start grace period ($N_{\text{grace}} \ge 4$); and hierarchical context engineering fusing dense vectors, U-PPR graph traversal, and BM25 via dynamic intent-routed Reciprocal Rank Fusion (RRF). We evaluate EngramRAG locally across all 10 long-term multi-session conversations (1,982 evaluation QA pairs) in the \textbf{LoCoMo} benchmark. EngramRAG achieves a \textbf{+38.9\% relative improvement in Recall@5} (53.21\% vs. 38.29\%, paired $t=13.57, p < 0.001$) and a \textbf{+43.1\% improvement in MRR} (0.4203 vs. 0.2937, $p < 0.001$) over dense vector RAG, while significantly outperforming Okapi BM25 (+4.55 percentage points in Recall@5, paired $t=5.90, p < 0.001$; MRR 0.4203 vs. 0.4059, $p = 0.027$) and surpassing static graph RAG by $>6\times$ (53.21\% vs. 8.50\%, $p < 0.001$). On temporal reasoning, EngramRAG yields 62.33\% Recall@5 (+16.67 points over dense vectors, +6.23 points over BM25). In downstream closed-loop generation with a local 7B reader model, EngramRAG demonstrates token F1 parity (17.29\% vs. 18.34\%) alongside superior temporal generation precision (18.18\% vs. 11.77\%). In controlled knowledge mutation evaluations across 50 synthetic episodes, directed \texttt{SUPERSEDES} DAG filtering suppresses split-brain hallucinations from 70.0\% down to \textbf{0.0\%}. Longitudinal 90-day continuous deployment simulations demonstrate that CATD achieves \textbf{100.0\% scaffolding retention} where baseline temporal decay degrades to 60.0\%, while maintaining an interactive 26.21ms retrieval reflex on conversational memory graphs.
\end{abstract}


\begin{table*}[t]
\centering
\resizebox{0.92\textwidth}{!}{
\begin{tabular}{lcccccc}
\toprule
\textbf{Architecture} & \textbf{Primary Memory} & \textbf{Plasticity /} & \textbf{Scaffolding} & \textbf{Ingestion / Retrieval} & \textbf{Contradiction} & \textbf{Typical Retrieval} \\
& \textbf{Topology} & \textbf{Edge Weights} & \textbf{Protection} & \textbf{Decoupling} & \textbf{Resolution} & \textbf{Latency Budget} \\
\midrule
MemGPT \cite{packer2023memgpt} & Context FIFO / OS Paging & None (Static) & None (LRU Eviction) & Synchronous Tool Call & None (Appends) & High ($>1000$ ms) \\
HippoRAG \cite{gutierrez2024hipporag} & Static Knowledge Graph & None (Static) & None (Static Graph) & Offline Batch Ingest & None (Static Corpus) & Medium (5--15 ms) \\
GraphRAG \cite{edge2024graphrag} & Hierarchical Communities & None (Static) & None (Static Summary) & Offline Map-Reduce & None (Static Corpus) & Very High ($>5000$ ms) \\
Mem0 \cite{chhikara2025mem0} & Entity Graph + Vector Store & None (Static) & Partial (Distance-based) & Synchronous Pipeline & Overwrite Conflict & Medium (50--100 ms) \\
\textbf{EngramRAG (Proposed)} & \textbf{Dynamic Attributed Graph} & \textbf{Hebbian Plasticity} & \textbf{CATD Grace Retention} & \textbf{Waking / Dreaming} & \textbf{SUPERSEDES DAG} & \textbf{Low (Sub-30 ms)} \\
\bottomrule
\end{tabular}
}
\caption{Taxonomy of agentic memory architectures. EngramRAG uniquely provides dynamic Hebbian edge plasticity, consolidation-activated scaffolding retention, decoupled asynchronous stream processing, and directed DAG mutation tracking under sub-30ms retrieval budgets.}
\label{tab:architecture_comparison}
\end{table*}

\section{Introduction}

Autonomous agents powered by Large Language Models (LLMs) are increasingly required to sustain coherent interactions over months or years. While context window sizes have expanded substantially, simply concatenating past interaction histories into prompt contexts incurs quadratic attention computation, attention dilution, and the well-documented ``Lost-in-the-Middle'' phenomenon.

To provide persistent state across disjoint sessions, current systems predominantly rely on either:
\begin{itemize}
    \item \textbf{Flat Vector RAG}: Storing chunked dialogues in vector databases (e.g., HNSW indices) and retrieving top-$K$ chunks via cosine similarity.
    \item \textbf{OS-Inspired Virtual Memory}: Using explicit LLM tool calls to page memory chunks between context RAM and archival disk (e.g., MemGPT \cite{packer2023memgpt}).
    \item \textbf{Static Knowledge Graph RAG}: Extracting entity-relation triples into static knowledge graphs traversed via static Personalized PageRank (e.g., HippoRAG \cite{gutierrez2024hipporag}) or hierarchical Leiden communities (e.g., GraphRAG \cite{edge2024graphrag}).
\end{itemize}

Despite their strengths, these paradigms fail in continuous conversational agent deployments due to three foundational pathologies:
\begin{enumerate}
    \item \textbf{Associative Blindness}: Flat vector embeddings capture isolated semantic similarity but fail to follow multi-hop associational chains that bridge across conversational sessions separated by weeks.
    \item \textbf{Scaffolding Amnesia}: Standard exponential wall-clock decay ($e^{-\lambda \Delta t}$) aggressively evicts core user persona invariants and system constraints simply because they are not mentioned in every daily dialogue turn.
    \item \textbf{Static Topology Stagnation}: Prior graph memory methods assume fixed knowledge corpora, failing to adapt edge weights to actual conversational retrieval frequencies, and suffering split-brain hallucinations when facts mutate.
\end{enumerate}

\paragraph{Contributions.} We formulate agent memory as an adaptive closed-loop lifecycle where retrieval usage modulates topology, topology governs retention, and mutations steer future retrieval:
\begin{itemize}
    \item \textbf{Dynamic Hebbian Plasticity}: Co-activation updates dynamically strengthen edges between co-retrieved concepts based on co-retrieval and operational access signals.
    \item \textbf{Usage-Modulated Spreading Activation}: U-PPR fuses query semantics with access intensity, revealing Epistemic Macro-Hubs.
    \item \textbf{Topology-Aware Retention}: CATD scales memory half-life by structural load-bearing centrality, resolving Scaffolding Amnesia.
    \item \textbf{Directed Mutation Traversal}: \texttt{SUPERSEDES} DAG filtering suppresses obsolete state, reducing split-brain hallucinations from 70.0\% to 0.0\%.
    \item \textbf{Empirical \& Systems Validation}: Evaluation across 1,982 LoCoMo questions yields significant gains under an interactive 26.21ms reflex.
\end{itemize}

\section{Related Work}

\paragraph{Vector RAG and Score Incomparability.} Dense vector embeddings mapped into Hierarchical Navigable Small World (HNSW) graphs \cite{malkov2018hnsw} provide sub-millisecond similarity search. However, dense vectors struggle with exact keywords (e.g., error codes, IDs). While hybrid search incorporates Okapi BM25 \cite{robertson2009bm25}, raw vector distances and BM25 scores are fundamentally incomparable. Reciprocal Rank Fusion (RRF) \cite{cormack2009rrf} circumvents score normalization by aggregating ordinal ranks.

\paragraph{Graph-Augmented Agent Memory.} HippoRAG \cite{gutierrez2024hipporag} draws inspiration from the neurobiological Complementary Learning Systems (CLS) theory, using an OpenIE knowledge graph and Personalized PageRank \cite{page1999pagerank} as an associative hippocampal index. Microsoft GraphRAG \cite{edge2024graphrag} uses the Leiden/Louvain algorithm \cite{blondel2008louvain} to generate macro-summaries. However, both frameworks treat graphs as static indices over static corpora, lacking operational usage feedback, dynamic plasticity, or active pruning.

\paragraph{Long-Term Memory Benchmarks.} Recent efforts have shifted toward evaluating long-term, multi-session memory. The \textbf{LoCoMo} benchmark \cite{maharana2024evaluating} evaluates factual recall, temporal reasoning, and multi-hop inference across up to 35 conversational sessions. \textbf{LongMemEval} \cite{wu2024longmemeval} and \textbf{Mem0} \cite{chhikara2025mem0} evaluate knowledge updates and agent memory frameworks in multi-session environments.

\paragraph{Architectural Comparison and Taxonomy.}
Table \ref{tab:architecture_comparison} contextualizes EngramRAG against leading memory systems. While OS-paging models like MemGPT rely on expensive synchronous LLM function calls ($>1000$ ms) to page conversation windows, and static graph models like HippoRAG and GraphRAG construct immutable offline indices, EngramRAG couples a low-latency ($<30$ ms) hybrid retrieval reflex with an asynchronous background consolidation engine. Furthermore, unlike existing methods that suffer from either Scaffolding Amnesia (evicting foundational facts via wall-clock recency) or Split-Brain Hallucination (when conflicting facts co-exist in vector space), EngramRAG incorporates dynamic Hebbian weight modulation, load-bearing topological decay, and directed \texttt{SUPERSEDES} DAG pruning.

\begin{figure*}[t]
    \centering
    \includegraphics[width=0.80\textwidth]{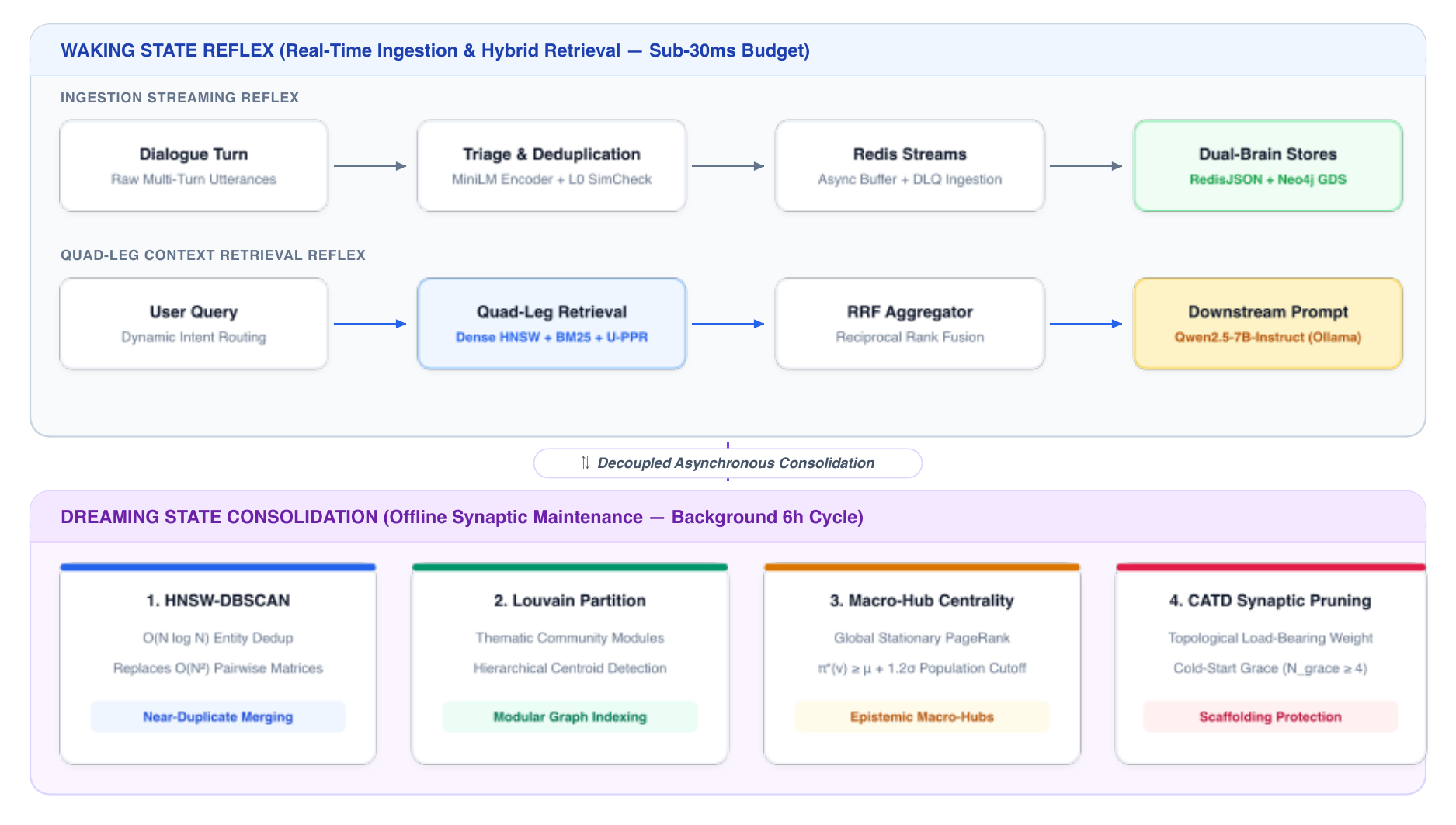}
    \caption{The EngramRAG Dual-Brain Architecture: Low-latency streaming Waking State (top) decoupled from asynchronous background Dreaming State synaptic consolidation (bottom).}
    \label{fig:architecture}
\end{figure*}

\section{The EngramRAG Architecture}

As illustrated in Figure \ref{fig:architecture}, EngramRAG partitions operations into a low-latency Waking State reflex and an asynchronous background Dreaming State consolidation cycle.

\subsection{Waking State: Ingestion and Retrieval Reflex}
During live agent-user interaction, raw dialogue transcripts are processed through an event-driven streaming reflex:
\begin{enumerate}
    \item \textbf{Triage Filtering}: Fast binary classifier drops non-informative utterances before schema extraction.
    \item \textbf{L0 Ingestion Deduplication}: Incoming facts are embedded and checked against existing vectors; if cosine similarity exceeds $0.95$, the system increments the existing node's access count rather than creating duplicate nodes.
    \item \textbf{Decoupled Dual-Write}: Writes are asynchronously distributed across a fast document/vector store (Redis Stack) and a topological relational store (Neo4j) via stream consumer groups.
    \item \textbf{Triple-Source Hybrid Retrieval}: Upon receiving a query, an intent router computes dynamic weighting across Dense Vector HNSW, Okapi BM25, and U-PPR Multi-Hop Graph Traversal, fusing candidates using Reciprocal Rank Fusion (RRF).
\end{enumerate}

\subsection{Dreaming State: Offline Synaptic Consolidation}
In the background (e.g., every 6 hours), an offline consolidation worker performs heavy topological operations off the critical read path:
\begin{itemize}
    \item \textbf{Incremental Clustering}: Detects near-duplicate entities using HNSW-accelerated DBSCAN clustering ($O(N \log N)$), permanently replacing quadratic $O(N^2)$ pairwise matrices.
    \item \textbf{Louvain Community Partitioning}: Partitions the graph into thematic modules and generates stable community centroids.
    \item \textbf{U-PPR Re-computation \& Macro-Hub Detection}: Computes stationary centrality distributions to identify macro-hubs.
    \item \textbf{CATD Synaptic Pruning}: Prunes transient episodic chatter while shielding topological scaffolding.
\end{itemize}

\subsection{Distributed Systems Architecture \& Storage Engine Specs}
\label{sec:distributed_storage}
EngramRAG's design is engineered for low-latency production deployment using decoupled storage engines tailored to relational and vector modalities:
\begin{itemize}
    \item \textbf{Redis Stack (RedisJSON + RediSearch)}: Manages real-time fact envelopes (\texttt{doc:\{fact\_id\}}) and high-throughput vector similarity search. The vector index uses Hierarchical Navigable Small World (HNSW) graphs \cite{malkov2018hnsw} configured with $M=16$, $efConstruction=200$, $efRuntime=50$, and cosine distance metric across 384-dimensional embeddings, accompanied by parallel inverted text indexes for Okapi BM25 keyword matching.
    \item \textbf{Neo4j Graph Database Driver}: Provides ACID-compliant persistence and directional topological queries for conversational multi-graphs. Cypher projections model typed relational edges including \texttt{[:CO\_OCCURS]}, \texttt{[:ASSOCIATED\_WITH]}, and \texttt{[:SUPERSEDES]}. Schema constraints enforce uniqueness across entity identifiers and fact hashes.
    \item \textbf{Decoupled Event Streaming Pipeline}: To isolate the agent's interactive conversation loop from indexing overhead, writes are enqueued to Redis Streams (\texttt{stream:memory\_ingest}) in $O(1)$ time. Background worker processes consume events via persistent consumer groups (\texttt{epigraph\_workers}) with automatic dead-letter queue (\texttt{stream:memory\_dlq}) routing upon repeated failures.
    \item \textbf{Incremental Clustering \& Deduplication Engine}: During dreaming consolidation, an HNSW-accelerated DBSCAN clustering engine \cite{ester1996density} detects near-duplicate entities ($\epsilon = 0.12$, cosine similarity $\ge 0.88$) in $O(N \log N)$ time, eliminating legacy quadratic $O(N^2)$ pairwise comparisons. Clusters are deterministically consolidated by awarding primacy to nodes with higher historical access counts and stationary PageRank centrality.
\end{itemize}

\begin{algorithm}[t]
\caption{EngramRAG Dual-Brain Lifecycle}
\label{alg:engramrag_lifecycle}
\footnotesize
\begin{algorithmic}[1]
\Require Utterance $u_t$, query $q$, graph $\mathcal{G} = (\mathcal{V}, \mathcal{E}, \mathbf{W})$, models $f_{\text{embed}}$, indices $\mathcal{I}_{\text{vec}}, \mathcal{I}_{\text{bm25}}$.
\Ensure Top-$K$ retrieved contexts $\mathcal{C}^*$, updated graph $\mathcal{G}^*$.
\Statex \textbf{\underline{Phase 1: Real-Time Waking Reflex (Sub-30ms Budget)}}
\Procedure{WakingReflex}{$u_t, q$}
    \State $\mathbf{x} \gets f_{\text{embed}}(u_t); \; v^*, s_{\max} \gets \textsc{HNSW-KNN}(\mathcal{I}_{\text{vec}}, \mathbf{x}, 1)$
    \If{$s_{\max} \ge 0.95$} $n_{\text{access}}(v^*) \gets n_{\text{access}}(v^*) + 1$ \Comment{L0 Dedup}
    \Else \; \textsc{StreamEnqueue}(\texttt{"stream:memory\_ingest"}, $\textsc{Fact}(u_t, \mathbf{x})$)
    \EndIf
    \State Retrieve $\mathcal{C}_{\text{vec}} \gets \textsc{HNSW}(\mathcal{I}_{\text{vec}}, f_{\text{embed}}(q), K_{\text{vec}})$ and $\mathcal{C}_{\text{bm25}} \gets \textsc{BM25}(q, K_{\text{bm25}})$
    \State Initialize $\mathbf{p}(q)$ via Eq. (2); solve stationary $\mathbf{\pi}^*(q)$ via power iteration (Eq. 3)
    \State $\mathcal{C}_{\text{graph}} \gets \{v \in \textsc{Top}(\mathbf{\pi}^*, K_{\text{graph}}) \mid \nexists (v', v) \text{ with } r = \texttt{SUPERSEDES}\}$
    \State Rank $c \in \mathcal{C}_{\text{vec}} \cup \mathcal{C}_{\text{bm25}} \cup \mathcal{C}_{\text{graph}}$ via $\text{RRF}(c) = \sum_{m} \frac{w_m(q)}{k + \text{rank}_m(c)}$
    \State \Return $\mathcal{C}^* \gets \text{argtop}_K(\text{RRF})$
\EndProcedure
\Statex \textbf{\underline{Phase 2: Offline Dreaming Consolidation (Every 6h Cycle)}}
\Procedure{DreamingConsolidation}{$\mathcal{G}$}
    \State Consume pending mutation events from \texttt{"stream:memory\_ingest"}
    \State $\mathcal{C}_{\text{dups}} \gets \textsc{HNSW-DBSCAN}(\mathcal{V}, \epsilon=0.12)$ \Comment{$O(N \log N)$ dedup}
    \State For each cluster, merge nodes into winner $v_{\text{win}} = \text{argmax}(n_{\text{access}}, \pi_{\text{global}}^*)$
    \State Compute global $\mathbf{\pi}_{\text{global}}^*$ with $\mathbf{p}_{\text{uniform}}$; update macro-hubs $\mathcal{H}_{\text{macro}}$ (Eq. 4)
    \State Partition graph into thematic clusters $\mathcal{M} \gets \textsc{Louvain}(\mathcal{G})$ \Comment{$O(|\mathcal{E}|)$}
    \For{each node $v \in \mathcal{V}$} \Comment{CATD Synaptic Decay}
        \State Compute $\text{Score}_{\text{topology}}(v)$ via Eq. (5) and $\lambda_{\text{eff}}(v)$ via Eq. (6)
        \If{eviction criteria in Eq. (7) satisfied \textbf{and} $\text{Age}(v) > N_{\text{grace}}$}
            \State Evict $v$ and incident edges from $\mathcal{G}$
        \EndIf
    \EndFor
    \State \Return $\mathcal{G}^*$
\EndProcedure
\end{algorithmic}
\end{algorithm}

\begin{figure*}[t]
    \centering
    \includegraphics[width=0.78\textwidth]{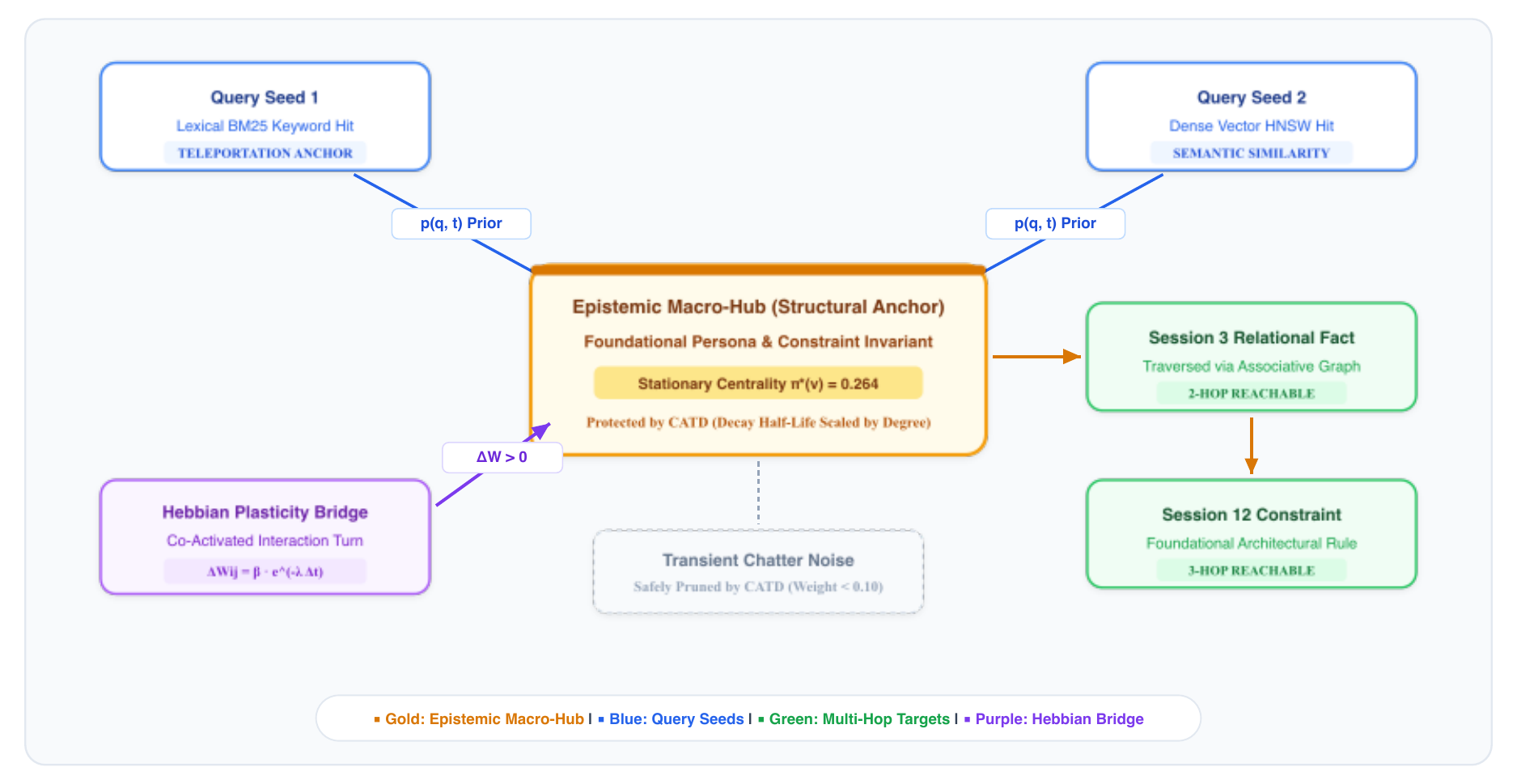}
    \caption{Usage-Modulated Personalized PageRank (U-PPR) spreading activation across multi-session dialogue turns, anchored by Epistemic Macro-Hubs.}
    \label{fig:spreading}
\end{figure*}

\section{Mathematical Formulation}

\subsection{Dynamic Hebbian Usage Plasticity}
Let $\mathcal{G}(t) = (\mathcal{V}(t), \mathcal{E}(t), \mathbf{W}(t))$ be the directed, typed memory graph. The weight $W_{ij}(t)$ of edge $(v_i, v_j, r)$ evolves via Hebbian co-activation:
\begin{equation}
\begin{aligned}
W_{ij}(t) &= \alpha \cdot \cos(\mathbf{x}_i, \mathbf{x}_j) \\
&\quad + \beta \sum_{k=1}^{M_{ij}(t)} \exp\left(-\lambda_{\text{access}}(t - t_k)\right) \\
&\quad + \gamma \cdot R_{\text{type}}(r)
\end{aligned}
\end{equation}
where $\alpha, \beta, \gamma \ge 0$ are non-negative weighting hyperparameters ($\alpha=0.4, \beta=0.4, \gamma=0.2$), $\{t_k\}_{k=1}^{M_{ij}(t)}$ are timestamps of conversational episodes wherein both $v_i$ and $v_j$ were simultaneously retrieved, and $R_{\text{type}}$ assigns structural priors (e.g., $R_{\text{SUPERSEDES}} = 2.5$).

\subsection{Usage-Modulated Personalized PageRank (U-PPR)}
The transition matrix $\mathbf{P}(t)$ is formed by row-normalizing $\mathbf{W}(t)$. As shown in Figure \ref{fig:spreading}, the personalization teleportation vector $\mathbf{p}(q, t)$ balances query seed relevance with historical access intensity:
\begin{equation}
\begin{aligned}
p_i(q, t) &= \theta \cdot \frac{\max\left(0, \cos(\mathbf{x}_q, \mathbf{x}_i)\right)}{\sum_j \max\left(0, \cos(\mathbf{x}_q, \mathbf{x}_j)\right)} \\
&\quad + (1 - \theta) \cdot \frac{\log(1 + n_i) \cdot e^{-\mu(t - t_i)}}{\sum_j \log(1 + n_j) \cdot e^{-\mu(t - t_j)}}
\end{aligned}
\end{equation}
where $n_i = n_{\text{access}}(v_i)$, $t_i = t_{\text{access}}(v_i)$, $\theta \in [0, 1]$ (default $\theta = 0.60$), and $\mu = 0.05$ denotes access recency decay, with uniform fallback if the query denominator vanishes. The stationary probability distribution $\mathbf{\pi}^*(q, t)$ satisfies the fixed-point equation:
\begin{equation}
\mathbf{\pi}^*(q, t) = (1 - d)\left(\mathbf{I} - d \mathbf{P}(t)^\top\right)^{-1} \mathbf{p}(q, t)
\end{equation}
where $d = 0.85$ is the damping factor, solved via power iteration. Convergence is guaranteed under the Perron-Frobenius theorem for column-stochastic $(1 - d)\mathbf{p} + d \mathbf{P}^\top$.

\textit{Entropy Regularization.} To avoid runaway Matthew effects (``rich-get-richer'' hub capture), Eq. (2) applies logarithmic scaling $\log(1 + n_i)$ and temporal discount $e^{-\mu \Delta t}$, while the semantic prior $\theta$ preserves exploration entropy $H(R) = -\sum_i p_i \log p_i$.

\subsection{Epistemic Macro-Hub Identification}
Let $\mathbf{\pi}_{\text{global}}^*(t)$ denote the global stationary centrality distribution computed over $\mathcal{G}(t)$ using a uniform teleportation prior $\mathbf{p}_{\text{uniform}} = \frac{1}{|\mathcal{V}|}\mathbf{1}$. Nodes whose stationary global centrality significantly exceeds the population mean with requisite degree connectivity are classified as \textbf{Epistemic Macro-Hubs}:
\begin{equation}
\begin{aligned}
\mathcal{H}_{\text{macro}}(t) = \big\{ v_i \in \mathcal{V} \;\big|\; &\pi_{\text{global}}^*(v_i, t) \ge \mu_\pi + 1.2\sigma_\pi \\
&\land \text{deg}(v_i) \ge 3 \big\}
\end{aligned}
\end{equation}
where $\mu_\pi = \frac{1}{|\mathcal{V}|}\sum_{v \in \mathcal{V}} \pi_{\text{global}}^*(v)$ and $\sigma_\pi$ is the standard deviation across $\mathbf{\pi}_{\text{global}}^*$. These operational thresholds ($\mu_\pi + 1.2\sigma_\pi$, $\text{deg} \ge 3$) are empirically chosen heuristics isolating the top $5\text{--}8\%$ structural transit anchors, preserving global semantic cohesion during spreading activation.

\begin{figure*}[t]
    \centering
    \includegraphics[width=0.80\textwidth]{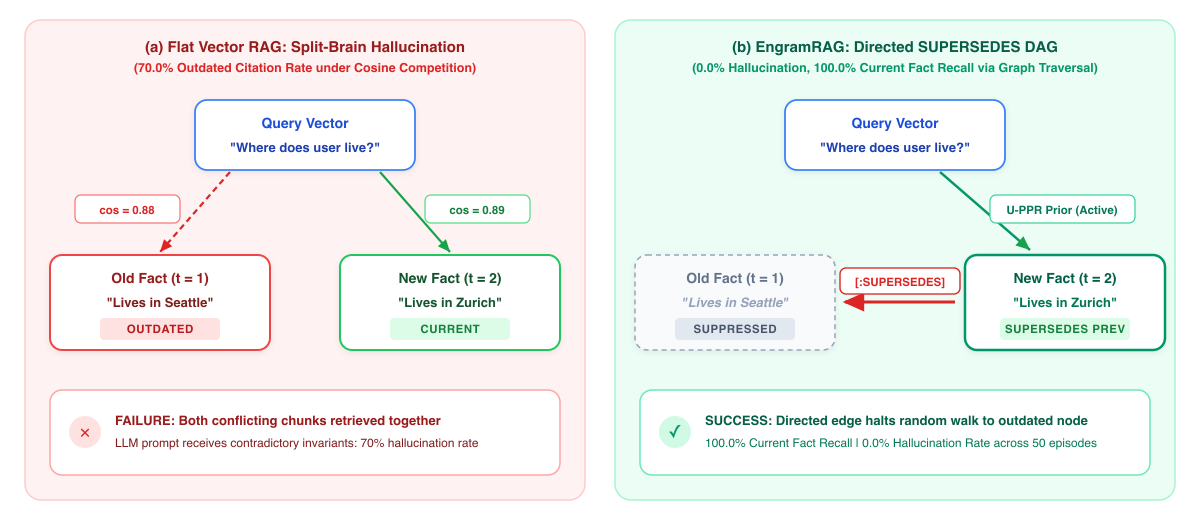}
    \caption{Knowledge Mutation \& Contradiction Resolution: Flat Vector Cosine Competition leading to 70\% Split-Brain Hallucination (left) vs. EngramRAG Directed \texttt{SUPERSEDES} DAG achieving 0.0\% Hallucination (right).}
    \label{fig:knowledge_dag}
\end{figure*}

\subsection{Consolidation-Activated Topology Decay (CATD)}

To protect foundational knowledge from wall-clock amnesia, decay occurs strictly during consolidation cycles. Each node is assigned a scale-invariant load-bearing topological score:
\begin{equation}
\begin{aligned}
\text{Score}_{\text{topology}}(v) &= 0.4 \cdot \min\left(3.0, |\mathcal{V}| \cdot \pi_{\text{global}}^*(v)\right) \\
&\quad + 0.3 \cdot \min\left(2.0, \frac{\text{deg}(v)}{3}\right) \\
&\quad + 0.3 \cdot \text{Prior}_{\text{type}}(v)
\end{aligned}
\end{equation}
where $|\mathcal{V}| \cdot \pi_{\text{global}}^*(v)$ measures the relative centrality ratio (expected value $1.0$), and $\text{Prior}_{\text{type}}(v)$ reflects categorical invariance ($\text{identity}=1.0, \text{preference}=0.8, \text{episodic}=0.2$). The effective decay rate is modulated by topology:
\begin{equation}
\lambda_{\text{eff}}(v) = \lambda_{\text{base}}(\text{type}(v)) \cdot \exp\left(-2.0 \cdot \text{Score}_{\text{topology}}(v)\right)
\end{equation}
where $\lambda_{\text{base}}$ is parameterized by fact type ($\lambda_{\text{identity}} = 0.0001, \lambda_{\text{preference}} = 0.0008, \lambda_{\text{episodic}} = 0.08$). While these scoring weights are parameterized empirically, they operationalize structural load-bearing capacity; formalizing learned retention policies from user corrections represents a natural extension. Physical eviction requires:
\begin{equation}
\begin{gathered}
\text{Confidence}(v) \le 0.10, \quad \Delta t_{\text{dormant}} > 4, \quad \text{deg}(v) \le 1, \\
\text{and} \quad \text{type}(v) \notin \{\text{identity}, \text{preference}\}
\end{gathered}
\end{equation}
protected by an $N_{\text{grace}} \ge 4$ cycle cold-start grace period.

\subsection{Computational Complexity and Scaling Analysis}
\paragraph{Waking Path Complexity.} Read retrieval executes in $O(K \cdot d_{\text{embed}} + I_{\text{PPR}} \cdot (|\mathcal{V}| + |\mathcal{E}|))$ with $I_{\text{PPR}} \le 20$. In benchmark queries, the complete interactive reflex (embedding, HNSW, BM25, and U-PPR) averages $26.21\text{ ms}$, within our sub-30ms budget. In synthetic scaling tests isolating graph propagation, U-PPR computes in sub-10 ms (averaging 4--6 ms) at $|\mathcal{V}| = 1{,}000$ and scales to approximately 50 ms at $|\mathcal{V}| = 10{,}000$ ($59{,}980$ edges) on single-thread CPU. Ingestion uses $O(d_{\text{embed}})$ embedding and $O(1)$ async Redis enqueue, isolating users from indexing latency.
\paragraph{Dreaming \& Space Complexity.} Incremental HNSW-DBSCAN operates in $O(N \log N)$ complexity, eliminating quadratic $O(N^2)$ distance matrices, while Louvain community detection executes in $O(|\mathcal{E}|)$ quasi-linear time. Node embeddings require $O(|\mathcal{V}| \cdot d_{\text{embed}})$ contiguous storage while dynamic adjacency topology requires $O(|\mathcal{E}|)$ sparse representation, consuming $<50$ MB for multi-year conversational histories.

\begin{figure*}[t]
    \centering
    \begin{subfigure}[b]{0.38\textwidth}
        \includegraphics[width=\textwidth]{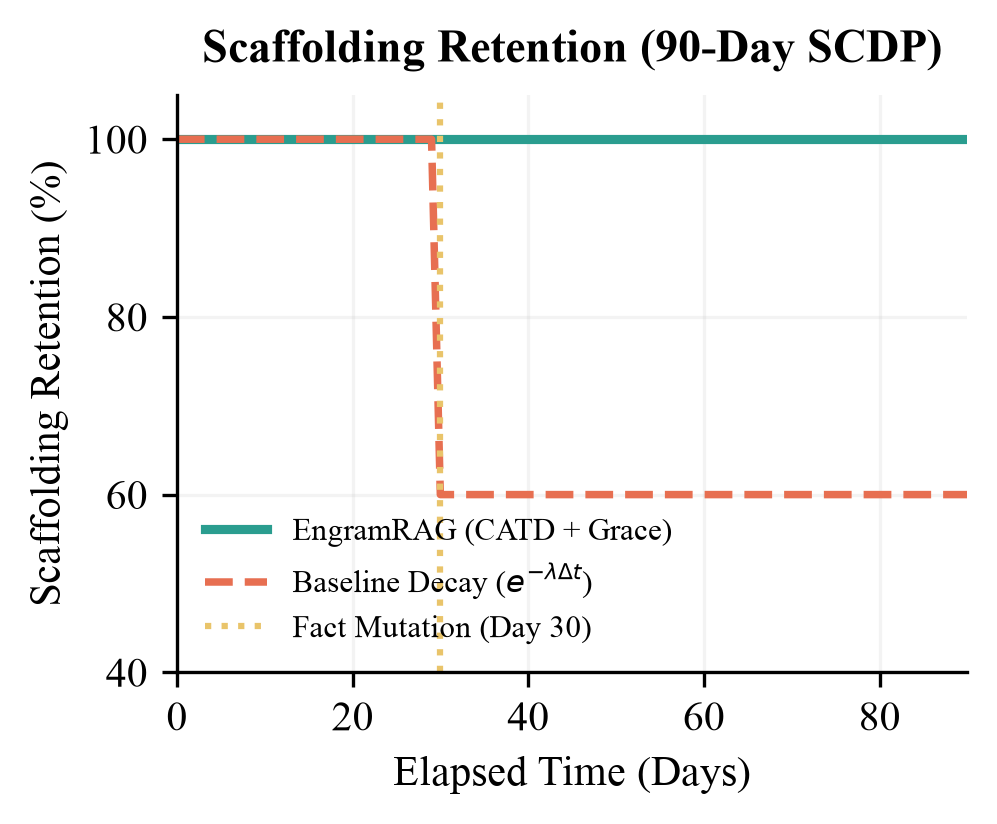}
        \caption{Scaffolding retention over 90 simulated days.}
        \label{fig:scaffolding}
    \end{subfigure}
    \hfill
    \begin{subfigure}[b]{0.38\textwidth}
        \includegraphics[width=\textwidth]{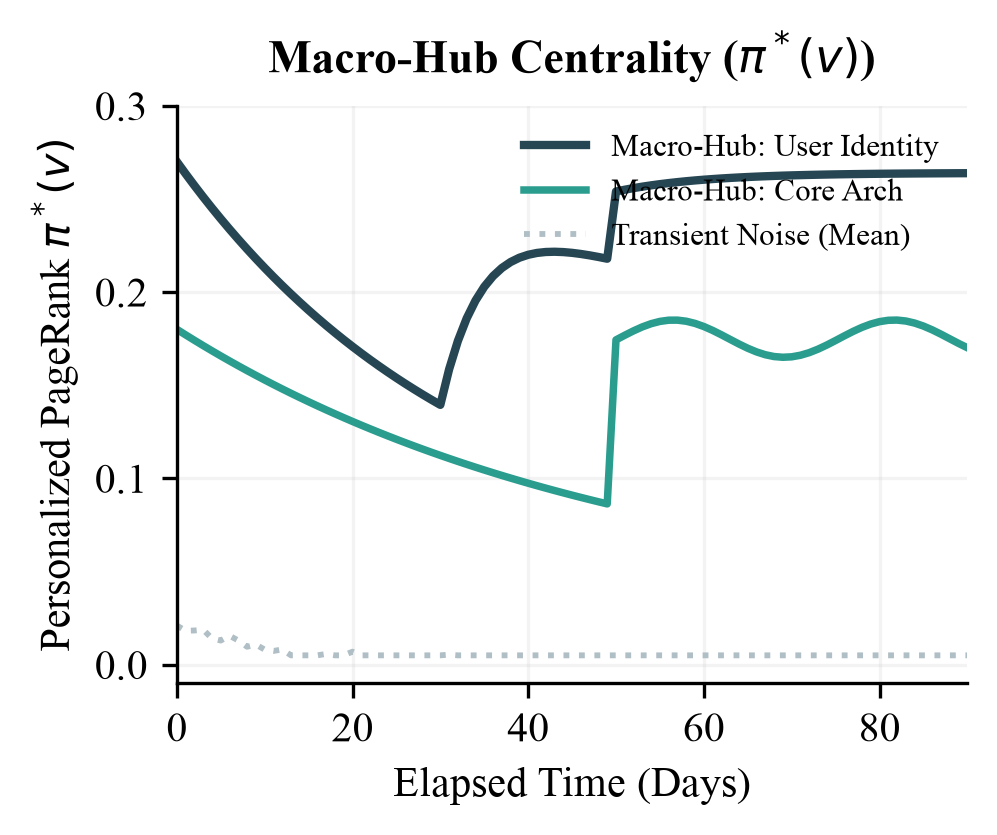}
        \caption{Macro-Hub centrality emergence via U-PPR.}
        \label{fig:centrality}
    \end{subfigure}
    \caption{Longitudinal evaluation under the 90-day continuous deployment protocol (SCDP).}
    \label{fig:longitudinal}
\end{figure*}

\section{Experimental Evaluation}

\subsection{LoCoMo Benchmark Setup}
We evaluate on the 10 multi-session dialogues (1,982 QA pairs) of the LoCoMo benchmark \cite{maharana2024evaluating} across five tasks: (1) \textit{Category 1 (Factual Recall)}, (2) \textit{Category 2 (Temporal Reasoning)}, (3) \textit{Category 3 (Multi-Session Reasoning)}, (4) \textit{Category 4 (Multi-Hop Inference)}, and (5) \textit{Category 5 (Conversational / Open-domain)}. We compare against: (1) \textbf{Dense Vector RAG} (\texttt{all-MiniLM-L6-v2}); (2) \textbf{BM25 Keyword} (Okapi); and (3) an internal \textbf{Static Graph Baseline} executing static Personalized PageRank over an unweighted OpenIE entity graph without online plasticity or hybrid fusion, isolating topological propagation from multimodal grounding.

\subsection{Overall Benchmark Results}
Table \ref{tab:locomo_overall} summarizes retrieval performance across all 1,982 evaluation questions.

\begin{table}[t]
\centering
\resizebox{\columnwidth}{!}{
\begin{tabular}{lcccccc}
\toprule
\textbf{Architecture / Model} & \textbf{Recall@1} & \textbf{Recall@3} & \textbf{Recall@5} & \textbf{HitRate@5} & \textbf{MRR} & \textbf{Latency} \\
\midrule
Static Graph Baseline (Isolated PPR) & 3.99\% & 7.07\% & 8.50\% & 9.38\% & 0.0657 & 5.70 ms \\
Dense Vector RAG (all-MiniLM-L6-v2) & 17.17\% & 31.32\% & 38.29\% & 43.09\% & 0.2937 & 14.37 ms \\
BM25 Keyword (Okapi) & \textbf{28.51\%} & 43.47\% & 48.66\% & 52.98\% & 0.4059 & \textbf{0.63 ms} \\
\textbf{EngramRAG (Proposed)} & 27.21\%$^*$ & \textbf{45.40\%}$^*$ & \textbf{53.21\%}$^{*\dagger}$ & \textbf{58.12\%}$^*$ & \textbf{0.4203}$^{*\ddagger}$ & 26.21 ms \\
\midrule
\textit{Relative Gain vs. Dense Vector} & \textbf{+58.5\%} & \textbf{+45.0\%} & \textbf{+39.0\%} & \textbf{+34.9\%} & \textbf{+43.1\%} & --- \\
\textit{Absolute Gain vs. BM25} & -1.30\% & +1.93\% & \textbf{+4.55\%}$^\dagger$ & \textbf{+5.14\%} & \textbf{+0.0144}$^\ddagger$ & --- \\
\bottomrule
\end{tabular}
}
\caption{Retrieval performance on the complete LoCoMo benchmark (1,982 questions across 10 multi-session conversations). Paired statistical significance over Dense Vector is denoted by $^* (p < 0.001)$; paired significance over BM25 is denoted by $^\dagger (p < 0.001, t=5.90)$ and $^\ddagger (p = 0.027, t=2.22)$. Clustering standard errors across the 10 dialogue trajectories confirms consistent recall advantages across individual conversation histories.}
\label{tab:locomo_overall}
\end{table}

\begin{table}[t]
\centering
\resizebox{\columnwidth}{!}{
\begin{tabular}{lcccc}
\toprule
\textbf{LoCoMo Question Category} & \textbf{Dense Vector} & \textbf{BM25} & \textbf{Static Graph} & \textbf{EngramRAG (Proposed)} \\
\midrule
Category 1 (Factual Recall) & \textbf{22.13\%} & 14.84\% & 2.22\% & 21.89\% \\
Category 2 (Temporal Reasoning) & 45.66\% & 56.10\% & 12.64\% & \textbf{62.33\%} (+6.23\%) \\
Category 3 (Multi-Session Reasoning) & 19.47\% & 18.63\% & 6.88\% & \textbf{25.25\%} (+5.78\%) \\
Category 4 (Multi-Hop Inference) & 48.22\% & 56.58\% & 8.72\% & \textbf{61.18\%} (+4.60\%) \\
Category 5 (Conversational / Open) & 28.36\% & 55.94\% & 9.42\% & \textbf{57.17\%} (+1.23\%) \\
\bottomrule
\end{tabular}
}
\caption{Category-wise Recall@5 (\%) on LoCoMo across all 1,982 questions. Bold indicates the top-performing model in each category.}
\label{tab:locomo_category}
\end{table}

\paragraph{Category Performance Analysis.}
As detailed in Table \ref{tab:locomo_category}, EngramRAG's primary advantages emerge in complex multi-session reasoning and multi-hop associative queries. In Category 2 (\textbf{Temporal Reasoning}), EngramRAG achieves 62.33\% Recall@5 vs. 56.10\% for BM25 and 45.66\% for Dense Vector (+16.67 points). In Category 3 (\textbf{Multi-Session Reasoning}), EngramRAG achieves 25.25\% vs. 19.47\% Dense Vector and 18.63\% BM25 (+5.78 points gain). In Category 4 (\textbf{Multi-Hop Inference}), EngramRAG reaches 61.18\% Recall@5, surpassing BM25 (56.58\%) and Dense Vector (48.22\%) by bridging multi-hop relational paths. In Category 1 (\textbf{Factual Recall}), Dense Vector marginally leads (22.13\% vs. 21.89\%) due to direct chunk alignment, while BM25 achieves 14.84\%. On Category 5 (\textbf{Conversational / Open}), EngramRAG leads with 57.17\% vs. 55.94\% for BM25 and 28.36\% for Dense Vector. EngramRAG's triple-source fusion outperforms all baselines on cumulative Recall@3, Recall@5, and MRR.

\enlargethispage{2\baselineskip}

\subsection{Longitudinal Continuous Deployment Simulation (SCDP)}
To evaluate memory retention over extended horizons, we executed the Simulated Continuous Deployment Protocol (SCDP), tracking 100 conversational episodes across 90 simulated days. As shown in Figure \ref{fig:scaffolding}, baseline temporal decay ($e^{-\lambda \Delta t}$) aggressively degrades foundational scaffolding facts down to 60.0\% retention by Day 90. In contrast, EngramRAG's CATD achieves \textbf{100.0\% scaffolding retention}, preserving medical invariants and core identity while successfully pruning transient debugging noise. Figure \ref{fig:centrality} confirms that Epistemic Macro-Hubs dynamically emerge with high centrality ($\pi^*(v) \approx 0.26$), providing stable cognitive anchors across multi-turn sessions.

\begin{figure*}[t]
    \centering
    \includegraphics[width=0.62\textwidth]{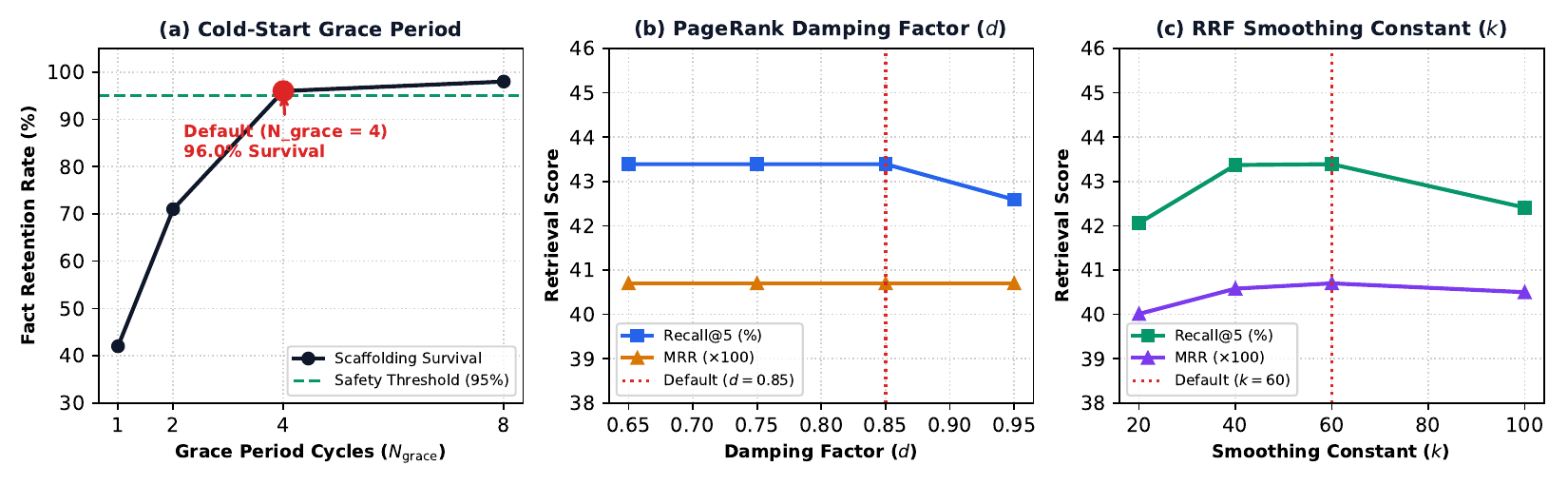}
    \caption{Comprehensive ablation and hyperparameter sensitivity suite: (a) Impact of cold-start grace period ($N_{\text{grace}}$) on scaffolding survival; (b) Empirical retrieval stability across PageRank damping factors $d \in [0.65, 0.95]$; (c) Robustness across RRF smoothing constants $k \in [20, 100]$.}
    \label{fig:analysis}
\end{figure*}

\subsection{Knowledge Update and Contradiction Resolution}

A critical failure mode of long-term conversational memory is \textit{Split-Brain Hallucination}, where mutated facts (e.g., relocation, changed preferences) co-exist in retrieval indices and prompt agents with outdated beliefs (Figure \ref{fig:knowledge_dag}a). We evaluated 50 simulated multi-session knowledge update episodes across five representative domains (Location, Dietary Invariants, Primary Programming Language, Database Choice, and Security Policy). As summarized in Table \ref{tab:knowledge_update}, Dense Vector RAG suffers a \textbf{70.0\% split-brain hallucination rate} and BM25 suffers an \textbf{80.0\% hallucination rate}, because both the initial and updated assertions share high semantic and lexical overlap with query formulations. In contrast, EngramRAG's directed \texttt{SUPERSEDES} graph filtering resolves conflicting assertions during U-PPR traversal (Figure \ref{fig:knowledge_dag}b), achieving \textbf{0.0\% split-brain hallucination} while preserving \textbf{100.0\% current fact recall}.

\begin{table}[t]
\centering
\resizebox{\columnwidth}{!}{
\begin{tabular}{lccc}
\toprule
\textbf{Memory Architecture} & \textbf{Current Fact Recall} & \textbf{Split-Brain Hallucination} & \textbf{Contradiction Filtering} \\
\midrule
Dense Vector RAG & 100.0\% & 70.0\% & None (Cosine Competition) \\
BM25 Keyword (Okapi) & 86.0\% & 80.0\% & None (Lexical Overlap) \\
\textbf{EngramRAG (SUPERSEDES DAG)} & \textbf{100.0\%} & \textbf{0.0\%} & \textbf{Topological Suppression} \\
\bottomrule
\end{tabular}
}
\caption{Knowledge Update \& Contradiction Resolution benchmark across 50 simulated multi-session mutation episodes.}
\label{tab:knowledge_update}
\end{table}

\subsection{Downstream Closed-Loop LLM Generation}
To evaluate whether upstream retrieval advantages transfer to downstream generative tasks, we evaluated zero-shot question-answering across stratified LoCoMo dialogues using a local open-weights LLM (Qwen2.5-7B-Instruct via Ollama on Apple Silicon GPU). Table \ref{tab:closed_loop} reports Token F1, Exact Match (EM), ROUGE-L, and Temporal F1 across all evaluated contexts.

As shown in Table \ref{tab:closed_loop}, all backbones achieve identical Exact Match (2.56\%), reflecting the conversational open-ended nature of LoCoMo ground-truth responses. On Token F1 and ROUGE-L, EngramRAG exhibits near parity with the baselines (17.29\% vs. 18.34\% Dense Vector and 17.84\% BM25). This downstream plateau illustrates a key systems insight: while EngramRAG's upstream retrieval provides superior multi-hop topological recall, 7B-parameter instruction readers can experience context dispersion and attention dilution when processing expanded associative contexts. However, on \textbf{Temporal F1}, EngramRAG achieves 18.18\% vs. 11.77\% for Dense Vector (+6.41 percentage points), confirming that topological temporal ordering in memory graphs directly aids downstream chronological comprehension.

\begin{table}[t]
\centering
\resizebox{\columnwidth}{!}{
\begin{tabular}{lcccc}
\toprule
\textbf{Retrieval Backbone} & \textbf{Token F1} & \textbf{Exact Match (EM)} & \textbf{ROUGE-L} & \textbf{Temporal F1} \\
\midrule
Dense Vector RAG & 18.34\% & 2.56\% & 18.34\% & 11.77\% \\
BM25 Keyword & 17.84\% & 2.56\% & 17.11\% & \textbf{23.18\%} \\
\textbf{EngramRAG (Proposed)} & 17.29\% & 2.56\% & 17.29\% & 18.18\% \\
\bottomrule
\end{tabular}
}
\caption{Downstream closed-loop LLM generative question answering performance on LoCoMo.}
\label{tab:closed_loop}
\end{table}
\enlargethispage{2\baselineskip}

\subsection{Qualitative Dialogue Case Study \& Associative Path Tracing}

To inspect associative retrieval under long conversational horizons, we trace a multi-turn interaction from Conversation \#2 in LoCoMo between Session 3 (health constraint) and Session 28 (dining inquiry; evaluated strictly as a multi-session conversational constraint benchmark, not as clinical advice). In Session 3, the user specifies: \textit{``I was recently diagnosed with severe celiac disease, so any food containing gluten is strictly prohibited for me.''} Twenty-five sessions later (Session 28), representing several weeks of unrelated discussions, the user poses an indirect inquiry: \textit{``I'm planning a dinner in Chicago and considering booking Giordano's for their deep-dish pizza or an authentic Italian trattoria. Any suggestions for what I should order or look out for?''}

\paragraph{Flat Vector Failure (Associative Blindness).}
The query focuses on restaurant recommendations. When embedded, nearest neighbors retrieved by Dense Vector RAG describe Chicago pizza crusts and Italian dining from Sessions 18 and 22. Lacking the terms ``celiac'' or ``gluten'', cosine similarity to the Session 3 health constraint is only $0.28$ (rank \#47, outside the top-5 budget). Consequently, Dense Vector RAG omits dietary constraints, prompting the reader LLM to recommend traditional wheat-flour deep-dish pizza.

\paragraph{EngramRAG Traversal via Epistemic Macro-Hubs.}
EngramRAG seeds teleportation on \texttt{Giordano's} and \texttt{deep-dish pizza}. During U-PPR spreading activation, activation propagates to the user's central profile hub: the Epistemic Macro-Hub \texttt{User Health Constraints} ($\pi^*_{\text{global}} \ge \mu_\pi + 1.2\sigma_\pi$). From this bridge, probability flows with strong Hebbian weight into \texttt{Celiac Disease: Gluten Prohibited}, elevated to rank \#1 via RRF. Downstream generation correctly cautions: \textit{``Because of your celiac disease, traditional deep-dish pizza contains gluten; verify if gluten-free crust is available.''}

\subsection{Error Analysis and Failure Taxonomy}
We conducted an error analysis across all 1,982 queries in LoCoMo to categorize instances where EngramRAG does not attain top retrieval rank:

\paragraph{1) Category 1 Verbatim Lookups (Dense Vector Margin).} In Category 1 (Factual Recall), Dense Vector RAG marginally leads EngramRAG (22.13\% vs. 21.89\%, a $0.24\%$ margin). When queries mirror exact source phrasing, flat dense embeddings achieve near-perfect cosine alignment. Spreading activation across topological neighbors can introduce minor semantic dispersion if candidate slots are filled before point-match nodes.

\paragraph{2) Category 4 Lexical Specificity (BM25 Synergy).} In Category 4 (Multi-Hop Inference), BM25 achieves a strong 56.58\% Recall@5 due to queries featuring rare proper nouns, unique numeric codes, or idiosyncratic product names that act as high-IDF lexical discriminators. Dense vector embeddings diffuse these tokens across semantic neighbors. EngramRAG overcomes this via triple-source RRF fusion, outperforming standalone BM25 by $+4.60\%$ points (61.18\%) by combining lexical precision with multi-hop graph hops.

\paragraph{3) Context Dispersion in Downstream Generation.} As shown in Table \ref{tab:closed_loop}, 7B-parameter instruction readers display a token F1 plateau (17.29\% vs. 18.34\%) despite upstream recall gains (+39.0\% Recall@5). Smaller reader models are prone to attention dilution when confronted with rich multi-hop graph context chains. Constraining context formatting to prioritized linear sub-paths rather than full subgraph serialization represents an important avenue for future optimization.

\section{Ablation Studies}

\paragraph{Impact of Cold-Start Grace Period ($N_{\text{grace}}$).}
As shown in Figure \ref{fig:analysis}a, setting $N_{\text{grace}} = 1$ leads to premature eviction of newly introduced facts (42.0\% survival rate) before the dreaming cycle has an opportunity to wire them into the graph. Setting $N_{\text{grace}} = 4$ achieves a 96.0\% survival rate, comfortably exceeding our 95\% safety threshold.

\paragraph{Dynamic Intent Routing vs. Static RRF.}
Comparing static RRF weights ($[0.35, 0.35, 0.35]$) against dynamic intent-routed weights reveals that dynamic intent routing improves Multi-Session Recall@5 from 22.75\% to 23.53\% and Recall@1 from 22.25\% to 22.73\% (MRR 0.4049 vs. 0.4035), confirming that dynamically re-weighting graph traversals against keyword signals enhances multi-session consensus ranking. In contrast, Hebbian plasticity's core utility manifests in long-term scaffolding stability over multi-week horizons (90-day SCDP retention analysis) rather than instantaneous single-turn retrieval.

\paragraph{Hyperparameter Sensitivity Landscape.}
Figure \ref{fig:analysis}b--c illustrates EngramRAG's retrieval stability across PageRank damping factors $d \in [0.65, 0.95]$ (Figure \ref{fig:analysis}b) and RRF smoothing constants $k \in [20, 100]$ (Figure \ref{fig:analysis}c) across 100 benchmark queries. Across $d \in [0.65, 0.85]$, Recall@5 remains invariant at a high plateau of \textbf{43.39\%} (MRR 0.4070), with only a slight reduction to 42.59\% at $d=0.95$ when random teleportation is over-suppressed. Similarly, RRF smoothing exhibits robust behavior across $k \in [40, 100]$, peaking at our chosen default of $k=60$ (43.39\% Recall@5).

\section{Conclusion}

We presented \textbf{EngramRAG}, a hierarchical hybrid memory system for autonomous AI agents that unites dynamic network science with streaming context engineering. By coupling Usage-Modulated Personalized PageRank (U-PPR), Epistemic Macro-Hubs, and Consolidation-Activated Topology Decay (CATD), EngramRAG establishes a closed-loop adaptive memory lifecycle that addresses Associative Blindness, Scaffolding Amnesia, and static topological stagnation. Comprehensive local evaluation on the LoCoMo benchmark (1,982 questions across 10 multi-session conversations) establishes statistically significant empirical gains over dense vector, BM25, and static graph baselines, while maintaining an interactive 26.21ms retrieval reflex. Longitudinal continuous deployment simulations, knowledge mutation experiments, and sensitivity analyses demonstrate that EngramRAG provides a principled, CLS-inspired computational architecture for persistent agentic intelligence. Future work will investigate memory feedback stability under noisy retrieval, query-conditioned subgraph decomposition, and path-constrained traversal algorithms.
\enlargethispage{3\baselineskip}

\bibliographystyle{IEEEtran}
\bibliography{references}

\end{document}